# Extending UML for Conceptual Modeling of Annotation of Medical Images

Mouhamed Gaith Ayadi
Department of computer sciences
ISG university of Tunis
Tunisia

Riadh Bouslimi
Department of computer sciences
ISG university of Tunis
Tunisia

Jalel Akaichi
Department of computer sciences
ISG university of Tunis
Tunisia

## ABSTRACT
Imaging has occupied a huge role in the management of patients, whether hospitalized or not. Depending on the patient's clinical problem, a variety of imaging modalities were available for use. This gave birth of the annotation of medical image process. The annotation is intended to image analysis and solve the problem of semantic gap. The reason for image annotation is due to increase in acquisition of images. Physicians and radiologists feel better while using annotation techniques for faster remedy in surgery and medicine due to the following reasons: giving details to the patients, searching the present and past records from the larger databases, and giving solutions to them in a faster and more accurate way. However, classical conceptual modeling does not incorporate the specificity of medical domain specially the annotation of medical image. The design phase is the most important activity in the successful building of annotation process. For this reason, we focus in this paper on presenting the conceptual modeling of the annotation of medical image by defining a new profile using the StarUML extensibility mechanism.

## 1. INTRODUCTION
We Advances in digital imaging technologies and the increasing prevalence of picture archival systems have led to an exponential growth in the number of images generated and stored in hospitals during recent years. Thus, automatic medical image annotation and categorization can be very useful for the purposes of image database management. In fact, the image is become probably one of the most important tools in medicine since it provides a method for diagnosis, monitoring drug treatment responses and disease management of patients with the advantage of being a very fast non-invasive procedure, having very few side effects and with an excellent cost-effect relationship. Doctors need to annotate these medical images and to analyze them to facilitate the access and taking care for the association of semantics to a medical image. We will work with annotation based CBIR. The goal of CBIR is to replicate this human ability of object recognition using a similar two step process: use of quantified measures from the image that are believed to represent color, shape, texture and interest points - the *image descriptors* - as an approach to human perception; use of *machine learning* techniques, to create a model for the data, or *similarity measures*, to interpret the image in order to establish the difference of two elements or groups of elements as an approach to human cognition. But the success of the annotation process rests on a good conceptual modeling schema. In fact, conceptual modeling offers a higher level of abstraction while describing the annotation process since it stays valid in case of technological evolution. However, no contribution is at the present time standard in term of data semantic models. This finding leads us to propose a new UML profile with user oriented graphical support to represent data and the modeling of annotation process with structural model (class diagram) and dynamic model (sequence diagram).

This paper is organized as follows. In section 2, we present a state of related to the medical image. In section 3, we present an overview of research works related to conceptual approaches and extensibility of UML for applications' needs. In section 4, we present the methodology that we adopted to extend the StarUML profile. In section 5, we present the UML profile. In section 6, we present the UML profile realization. In section 7, we summarize the work and we propose some perspectives that can be done in the future.

## 2. STATE OF THE ART
In this section, we try to present all the concepts related to the annotation of medical images.

*A. Annotation of Medical Image*

The annotation of medical images is the task of assigning each image a set of tags or keywords. The annotation can be defined differently according to Barnard et al **[1]** "The purpose of the annotation is to assign keywords to specific regions of the image." According authors **[2]** "they are learned attributes to describe objects. "Wang et al **[3]** and Weston et al **[4]** order the most similar to ours must define a "description" of the image that could help visual search.

*B. Annotation types*

There are generally three types of image annotation: manual, semi-automatic and automatic. The **manual annotation** is usually performed by a librarian named iconographer. Its role is to assign each image categories and groups of words, often taken from a thesaurus to find images easily. However, when you have a large volume of images to annotate this work quickly becomes tedious, if not impossible; this is not the case for automatic annotation.

The **automatic annotation**: This is the task of associating a set of words in an image using a computer system without human intervention. The advantage of this system is that it gives the possibility to the user to pose queries in a high level language allowing him to express his information need easily. But sometimes it is inefficient and poorly annotated images. To make a compromise between these two tasks, the combination has become necessary. This is what is known as the "**semi-automatic annotation**."

*C. Image Descriptor*

- *The histogram*

The histogram **[5]** shows the proportion of pixels of each color in an image. On the question, the user has two options: either to specify the percentage of each desired color or to propose a model picture.

- *The texture descriptors*

The texture **[6]** is the difference between the pixels in an image. In fact, this kind of descriptor is used when it comes to measure the similarity between the regions of two images of the same color. The interrogation by the texture can be done in the same way as the color (selected examples of textures, image presentation models). It is based on various methods such as: co-occurrence matrix, the Fourier transform of the Gabor filter, wavelet...

- *Shape descriptors*

This type allows you to compare two images with objects of the same shape **[7].** It offers the user to formulate his query using either a model image or predefined shapes compared to all images stored in the database. The shape features are often extracted from the segmentation of images.

- *SIFT*

The descriptor SIFT (Scale Invariant Feature Transform) **[5]** describes the neighborhood of a point by building a histogram of gradient orientations. The gradient vector (direction and amplitude) is calculated for each pixel scale corresponding to the neighborhood. Directions are discretized to have only more than 8 directions. The neighborhood is divided into a grid.

- *SURF (speeded Up Robust Features)*

Also said robust feature is an accelerated algorithm for detecting characteristic presented by researchers at the ETH Zurich and the Katholieke Universiteit Leuven. Its main objective is to accelerate the various image processing. SURF **[5]** is partly inspired by the SIFT descriptor, it outperforms fast and is more robust for different image transformations.

*D. Image Retrieval models*

According to **[8],** we identify three types of indexing models:

- the Boolean model
- the vector model
- And the probabilistic model.

For The Boolean model, the images are first characterized by a list of descriptors and the query is a logical formula that combines descriptors examples and logical operators (AND, OR, NOT). In response, the system is classified into two classes corresponding firstly to images that match the query and also to those who do not comply.

In the case of vector model, the query image and target image, that is to say, the images of the database are represented by a vector in a space attribute. This vector corresponds to the concatenation of the basic weights of descriptors. A function of similarity between vectors is used to classify the images according to their relevance to the query. For the probabilistic model, there assigning a relevance probability of the image in response to the request, each of the descriptors. Under the hypothesis of independence of the descriptors, it is possible to calculate the probability that the image meets the user's query as the product of the previous probabilities.

This type of model uses a strong user interaction through, for example, of relevance judgments issued by the user on the proposals made by the system results.

*E. CBIR Content-Based Image Retrieval*

In practice the conceptualization of a general thesaurus of medical terms consume many resources and demands extensive collaboration efforts where consensus is hard to reach. It is reasonable to use inductive approaches by starting with more specific standards and attempt generalization later. In the composite SNOMED-DICOM micro-glossary **[9]** such a strategy is used. Nevertheless, all standards presented are not ineffectual since they are used in several Picture Archive and Communications Systems (PACS). Facing the amount of images in a database, annotation by human hand can be a time consuming and cumbersome task where perception subjectivity can lead to unrecoverable errors. A study of medical images using DICOM headers revealed 15% of annotation errors from both human and machine origin **[10].** The amount of different languages that can be used for annotation is extensive and may lead to translation/interpretation errors during a search statement or when databases are merged. It is convenient to be aware of the prospect of re-indexing images due to the presence of an event that changes the importance of a particular aspect, e.g., Forsyth's previously unknown famous person photos **[11],** or the need to link the content of the image to a new search statement possibility, e.g., Seloff's engineer search for a misaligned mounting bracket existent only in a annotated astronaut training image **[12]**. Another major obstacles for concept-based image retrieval systems are the existence of homographs and the fact that the search statement, or *query*, does not allow the user to switch and/or combine interaction paradigms **[13]** during text transactions. The ideal system would relieve the human factor from the annotation task, by doing it automatically, and allowing image retrieval by its content in its purest form, not only by text description. This is Content Based Image Retrieval (CBIR).

## 3. RELATED WORK

One of the current concerns in software development is to better understand the domain of the problem, about which it is intended to create solutions that meet satisfactorily the real needs of users. To aid in this task, one of the techniques used is the conceptual modeling, which consists in to extract from the real world only those essential elements observed, leaving out implementation aspects. In this section, we present different approaches related to the conceptual modeling methodology, then we present research works that extended UML to adopt it to their conceptual modeling needs. The process of conceptual modeling allows a better understanding of the system being designed and is performed with the aid of specific modeling languages, which are languages whose syntax and semantics are focused toward the conceptual representation of a system **[14].** The Unified Modeling Language (UML) has been widely used and accepted by the scientific community and industry, as a tool for design and specification of systems **[15].**

One of the most important concerns when elaborating a *Model-Driven Development* (MDD) solution **[16]** is the specification of a modeling language that allows the required software products to be represented at the conceptual level without ambiguity. Among the different choices that exist for the definition of an adequate modeling language, there are two alternatives that appear to be the most suitable. The first of these is the creation of a specific language that is tailor-made for the MDD approach. This language is called *Domain-Specific Modeling Language* (DSML) **[17].** The second alternative is the customization of UML by means of extensions defined in the UML meta-model, which represent the abstract syntax related to the semantics required for the MDD proposal **[18].**

The second alternative for the specification of an adequate modeling language is the use of UML, which is a widely known modeling language that has a lot of support tools. However, it is also true that the semantics of certain UML conceptual constructs does not provide enough precision for its application in effective MDD processes. This lack of precision can be easily perceived in the Semantic Extension Points defined in the UML specification **[19],** where different semantic representations for one conceptual construct or the lack of definition of the appropriate semantics can be found.

An example of this is on of the semantic extension points related to the UML association, which state that "The order and way in which part instances in a composite are created is not defined" **[19].** In order to introduce the required semantic precision into UML, there are different extension mechanisms that can be used **[20],** One of them, the UML Profile extension mechanism, is the most suitable extension alternative because it is part of the UML standard and, hence, the extensions defined with a UML profile can be supported by UML-based tools. Therefore, existent MDD technologies based on UML (such as requirement traceability tools or cost estimation tools) can be reused by other MDD solutions that also use UML as base modeling language. In addition the UML profile extension mechanism allows the existent UML editors to be used thereby reducing the costs of implementing specific model editors.

However, since UML is a general purpose modeling language, the modeling facilities that the existent UML editors provide may not be the most appropriate to perform specific modeling tasks related to MDD approaches. Furthermore, the extension capabilities of the UML profile present limitations that in some cases might prevent a correct representation of all the modeling needs that are required by MDD proposals. After analyzing these two modeling alternatives (UML and DSMLs), an interesting modeling approach would be to provide a hybrid modeling schema that integrates both alternatives. This integration can be obtained by means of the transparent interchange of UML models and DSML models. Thus, it would be possible to take advantage of the existent UML tools for those models that can be represented by means of UML and only to implement specific tools for those models that require more complex modeling capabilities. It would also be possible to implement specific DSML-based tools for those features that are outside of the scope of existent UML tools.

The proposal presented in [**21**] defines a solution to solve these structural differences in order to obtain an adequate input for an automated UML profile generation. In addition, considering that the UML profile is generated from the DSML meta-model, during the generation of the UML profile also can be obtained the information of the equivalences (mapping) between the extended UML meta-model (extended with the generated UML profile) and the DSML meta-model.

Conallen **[22]** proposes a UML profile named Web Application Extensions (WAE). WAE extends UML to provide Web specific constructs for modeling WISs, including a new model called User Experience (UX) Model, which defines guidelines for modeling layout and navigation. OOWS (Object Oriented Web Solution) **[23]** uses UML for most of its models, making use of its extension mechanisms. But it also proposes extensions that are not standard, which can make things difficult for developers that do not have CASE tools specifically designed for the method.

Authors **[24]** propose an UML profile developed specifically for conceptual modeling of geographic databases called GeoProfile. This is not a definite proposal; they view this work as the first step towards the unification of the various existing models, aiming primarily at semantic interoperability.

## 4. ADOPTED METHODOLOGY

For the abstraction levels **[25]** (conceptual, logical and physical) our solution is established to cover the conceptual level. Here is a plan showing the position of our solution:

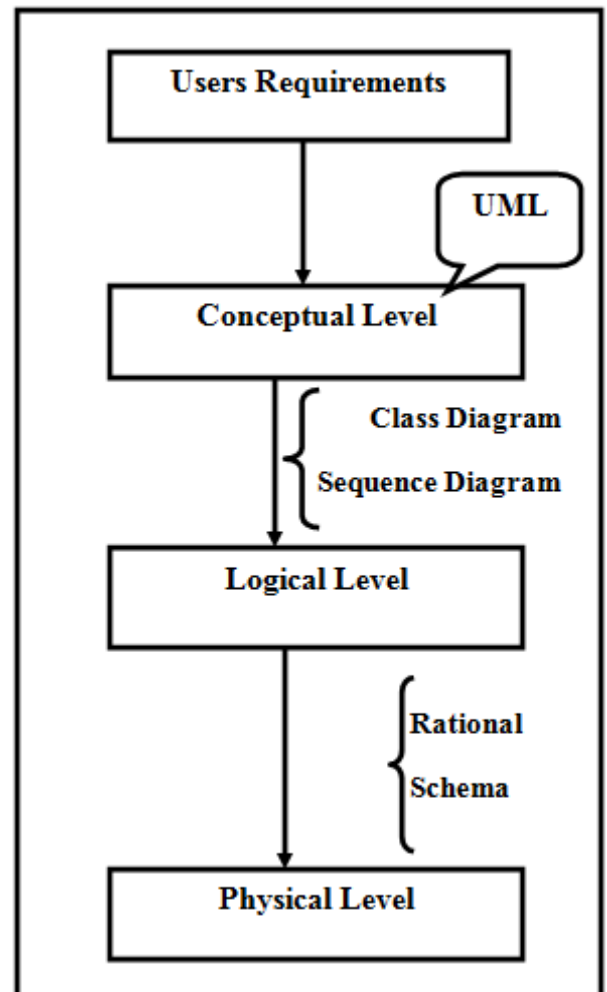

We chose to adopt the annotation process which is based on the UML profiles.

## 5. UML PROFILE

Despite being a general purpose language, which can be used in different application domains, there are situations in which the UML elements are not able to express all the peculiarities of a given domain. Therefore, to prevent the UML became too complex, it was specified as an extensible language **[26].** The OMG defines two ways of extending the UML. The first is based on the modification of the UML meta-model, thereby creating a new language, in which the syntax and semantics of the new elements are adapted to the intended domain. The second way is to adapt the UML to specific domains or platforms using the mechanism of profiles. In this second alternative, the elements of language are specialized, but respecting the UML meta-model and maintaining the original semantics of the elements unchanged **[27].**

In this first form of UML extension, the new language is created using MOF. In the second alternative, the language elements will be specialized by using the extension mechanisms provided by UML, which are:

- **Stereotypes**: A stereotype defines how an existing meta-class may be extended and enables the use of specific terminology for a domain or different platform in place of or in addition to the terminology used for the extended meta-class. Stereotypes can also change the appearance of the elements of the extended model using graphic icons;

- **Tagged values**: They are additional meta-attributes associated with a meta-class of the meta-model extended by a profile and add information to elements of the model;

- **Constraints**: These are restrictions associated with the corresponding elements of the metamodel. They can be written using natural language or OCL, which is also standardized by the OMG.

An UML profile is a set of extension mechanisms grouped in an UML package stereotyped as <<profile>>. As mentioned earlier, these mechanisms allow the extension of the syntax and semantics of the UML elements, but without violating the original semantics of UML and, therefore, consistent with MOF.

The idea of extending the UML for specific purposes is not new. UML 1.1 could already easily assign stereotypes and tagged values to model elements. However, the notion of profile was defined to provide a more structured and precise extension **[28].** UML profile is already adopted as a standard modeling in some domains, such as CORBA architecture **[29].** Other profiles are in the process of being adopted by the OMG or are being created by private organizations, software companies and research centers.

OMG [18] emphasized that there is no simple answer to the question of when to create a new metamodel or when to use the mechanism of profiles. Each alternative has its advantages and disadvantages, but the use of UML profiles provides a better cost-benefit ratio, by utilizing the entire structure of the UML tools and training materials. Fuentes and Vallecillo **[27]** mention that the benefits of using UML profiles undoubtedly exceed their limitations.

An UML Profile allows a structured and precise extension of UML constructors to customize UML for a particular domain. A well-specified UML Profile will have direct support of CASE tools. In other words, once the Profile is defined there is no need to implement new CASE tools. *Enterprise Architect* [9] and *Rational Software Modeler* **[21]** are examples of CASE tools with support for UML Profiles.

Hence, the development of a UML Profile has proven an excellent method to standardize modeling of specific domains, as it uses the language's popularity and tools compatible with UML 2.0, favoring standard acceptance and reducing time for training in new languages.

An UML profile **[30]** allows specializing UML in a precise domain, it consists of stereotypes, tagged values and constraints. A stereotype **[27]** is an element of the model that defines new values, new constraints and a new graphic representation. Its role is to give a semantic representation to an element of the model. A stereotype can be represented as a string character between two quotation marks << >> or with an icon. A marked value specifies a new property attached to an element of the model. It is represented between {} and placed with the name of another element. A constraint can become attached to any element of the model to refine its semantics and prevent an arbitrary use of the various elements.

It can be defined with the natural language and\or with the OCL (object constraint language) **[30]** which is a declarative language that allows developers to write constraints on the model's objects. Recently, UML profiles have a great progress in the ways for conception of Annotation Process. We present in this section, a conceptual solution for data warehouses design. We proceeded by an UML profile in order to add stereotypes. Our UML profile contains the Class Diagram and Sequence Diagram.

## 5.1 Class Diagram

We defined in this diagram stereotypes and icons related to the annotation of medical image. This diagram can be used in each case based on annotation.

*Classes Stereotypes*: We defined in this table classes's stereotypes used in the modeling.

**TABLE I**
**STEREOTYPES DESCRIPTION AND REPRESENTATION**

| Stereotypes name | Class type | Description | Icon |
|---|---|---|---|
| << Annotation >> | class | This stereotype indicates that the class represents the annotation | 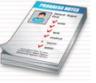 |
| << Medical image >> | class | This stereotype indicates that the class represents the image for the annotation | 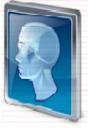 |
| << Annotator >> | class | This stereotype indicates that the class represents the annotator of medical image | 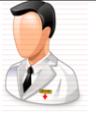 |
| << Patient >> | class | This stereotype indicates that the class represents the patient | 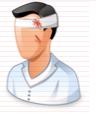 |
| << Key Words >> | class | This stereotype indicates that the class represents the key words of annotation | 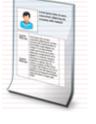 |

At the end of this section, we propose the following example of defining an UML Class related to the Class Diagram with an extended stereotype and icon using XML in the figure below.

### 5.3 UML profile realization

To implement our approach we chose the StarUML open source platform that uses the language XML to create the profiles UML. In this section we describe StarUML by showing its stretchable parts, and then we model a DW and their components with our UML profile.

#### 5.3.1 The StarUML profile

StarUML is a modeling platform with the UML language, conceived to support the MDA (Model Driven Architecture) approach.

### 5.2 Sequence Diagram

For sequence diagram we kept the standard elements of UML. Besides, we added some icons that can exist between different objects of Annotation as described in the following table:

**TABLE III**
**STEREOTYPES OF SEQUENCE DIAGRAM**

| Stereotypes name | Class type | Description | Icon |
|---|---|---|---|
| << Interface >> | Object | This stereotype indicates that the Interface. | |
| << Controler >> | Object | This stereotype indicates that the controller. | |

In this section, we present the task which is done by the annotator. The annotator send the query image and consult the table "Image" for searching the similar image. The system calculates the distance between query image and other images. In fact, the system checks the images and displays all images with their descriptions. Finally the annotator makes the annotation by choosing the similar image.

We propose the following sequence diagrams that illustrate the description cited above.

It is characterized by a strong flexibility and an excellent extensibility of its features. Indeed, besides the predefined functions, StarUML allows the addition of new functions which can be adapted to the user's needs. The inconveniences of this platform are that it does not allow specifying more than a stereotype for an element and it excludes the definition of the constraints. Thus in our work we considered that every element has only a single stereotype*.*

#### 5.3.2 The implementation of Annotation-UML profile

An UML profile is one package belonging to the mechanism of extension. This package is stereotypical < < Profile > > which is written in XML as we see in the following figure:
In the StarUML platform, we added a profile UML called "Medical Image Annotation" that contains "class diagram".

Indeed, we have created a file XML for the profile. Inside this file we appealed to extensions of notation which allows realizing specific notations that are different from those contained in UML.

```xml
<?xml version="1.0" encoding="UTF-8"?> <PROFILE version="1.0">
<HEADER> <NAME>MedicalImageAnnotation</NAME>
<DISPLAYNAME>MedicalImageAnnotation</DISPLAYNAME>
<DESCRIPTION>Medical Image Annotation conceptual modeling</DESCRIPTION>
</HEADER>
```

In this part, we represent the interfaces of our added profile in the figure below.

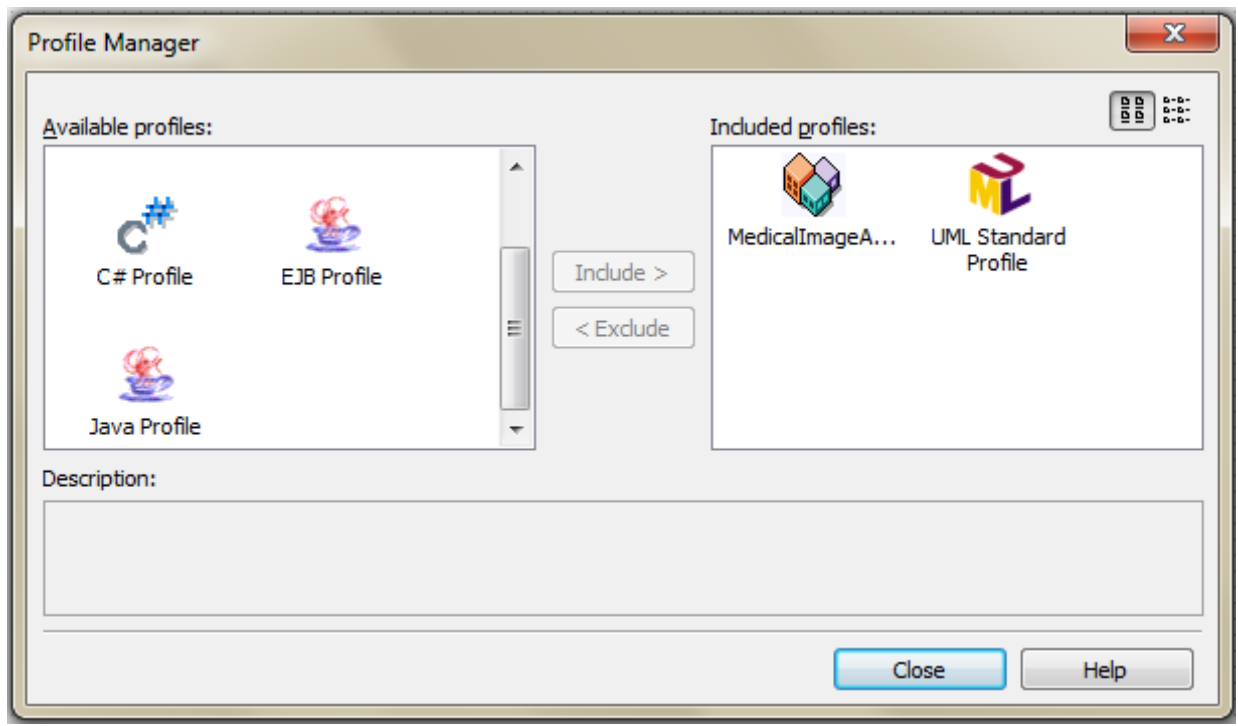

**Fig 1. New Image Annotation profile**

Now, on the tab **Model Explorer** of upper right select the object "**Untitled**" with a click of the mouse right button and choose the option: **Add-> Design Model** to create a new blank drawing template.

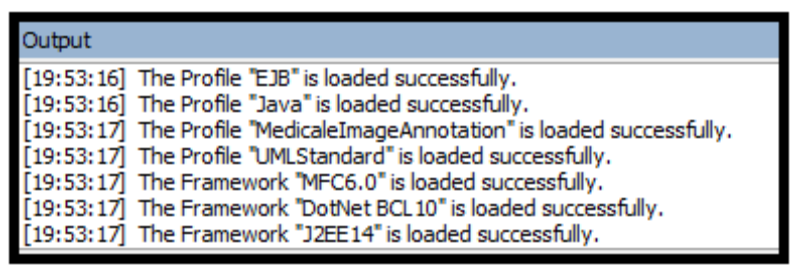

**Fig 2. Loading profit**

It is now possible to apply the *Annotation* stereotypes in UML elements of the diagram you created earlier. To test the *Annotation Profile***,** add an element **Medical Image Annotation Class** on the diagram. Having the class selected, in the tab **Properties** go into Stereotype.

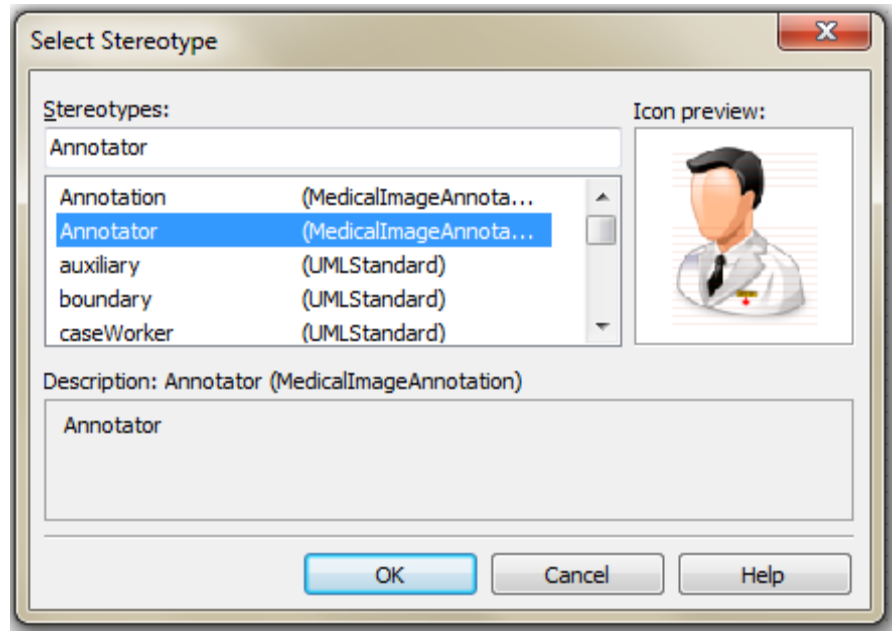

**Fig 3. Stereotype of new image annotation**

Note that, when you select a stereotype of *Annotation*, its icon will be shown on the field **Icon Preview**. Select any stereotype and click **OK** to apply it to the class. Note that the name of the stereotype will be displayed between <<...>> at the top of the class. *StarUML* only allows us to add one stereotype per element and allows you to show stereotypes in textual form "Textual", in form of icon "Iconic" or both "Decoration". To change the preview type, select the object and, in the top tab, select the type of view you want in the option **Stereotype Display**. Viewing the stereotype can also be changed by right-clicking on the object in the option **Format option-> Stereotype Display**.

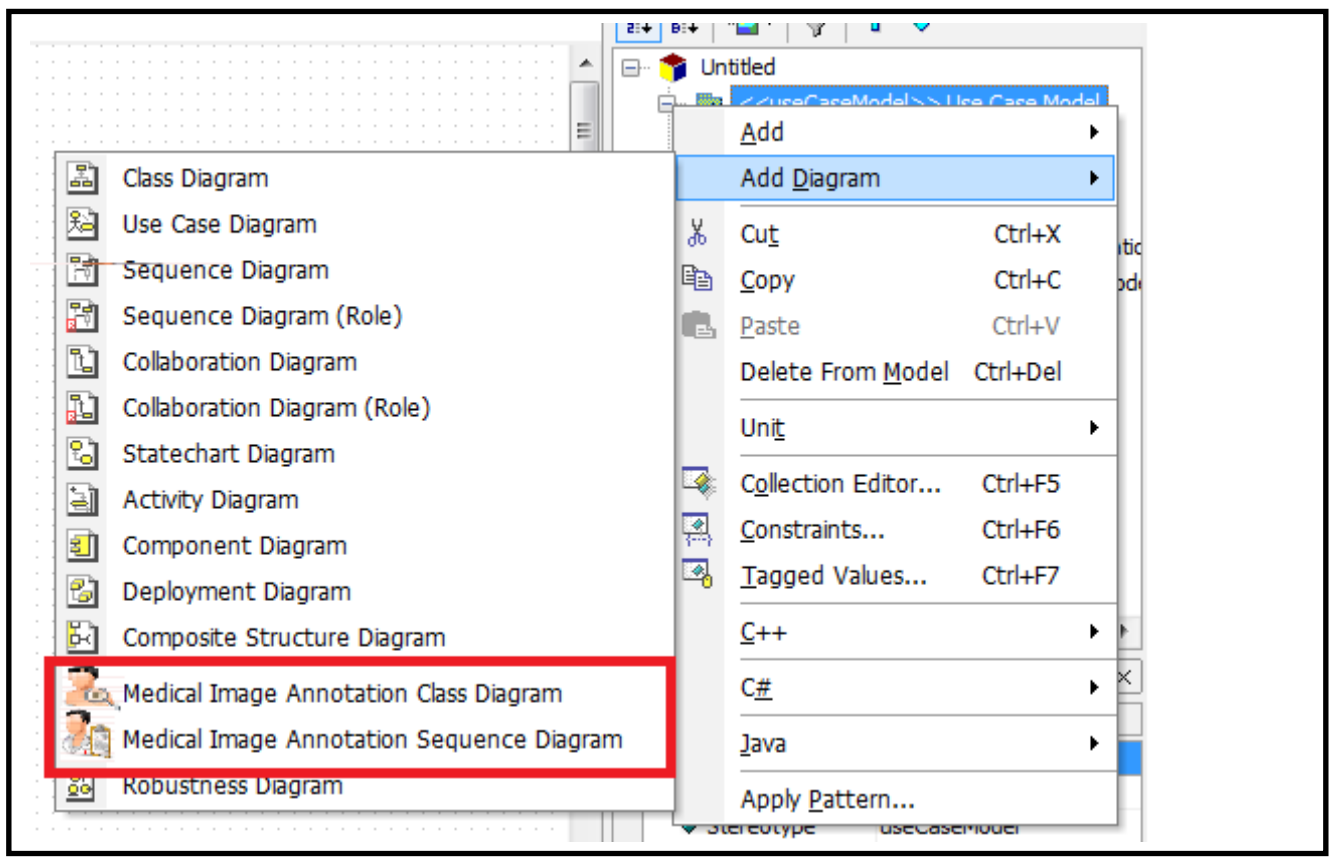

**Fig 4.Diagrams of Annotation of Medical Image**

We created a new Class Diagram, in which we added some stereotypes to identify each class (entity). In this diagram, there are some stereotypes and icons that can be used in any application related to the annotation of medical image.

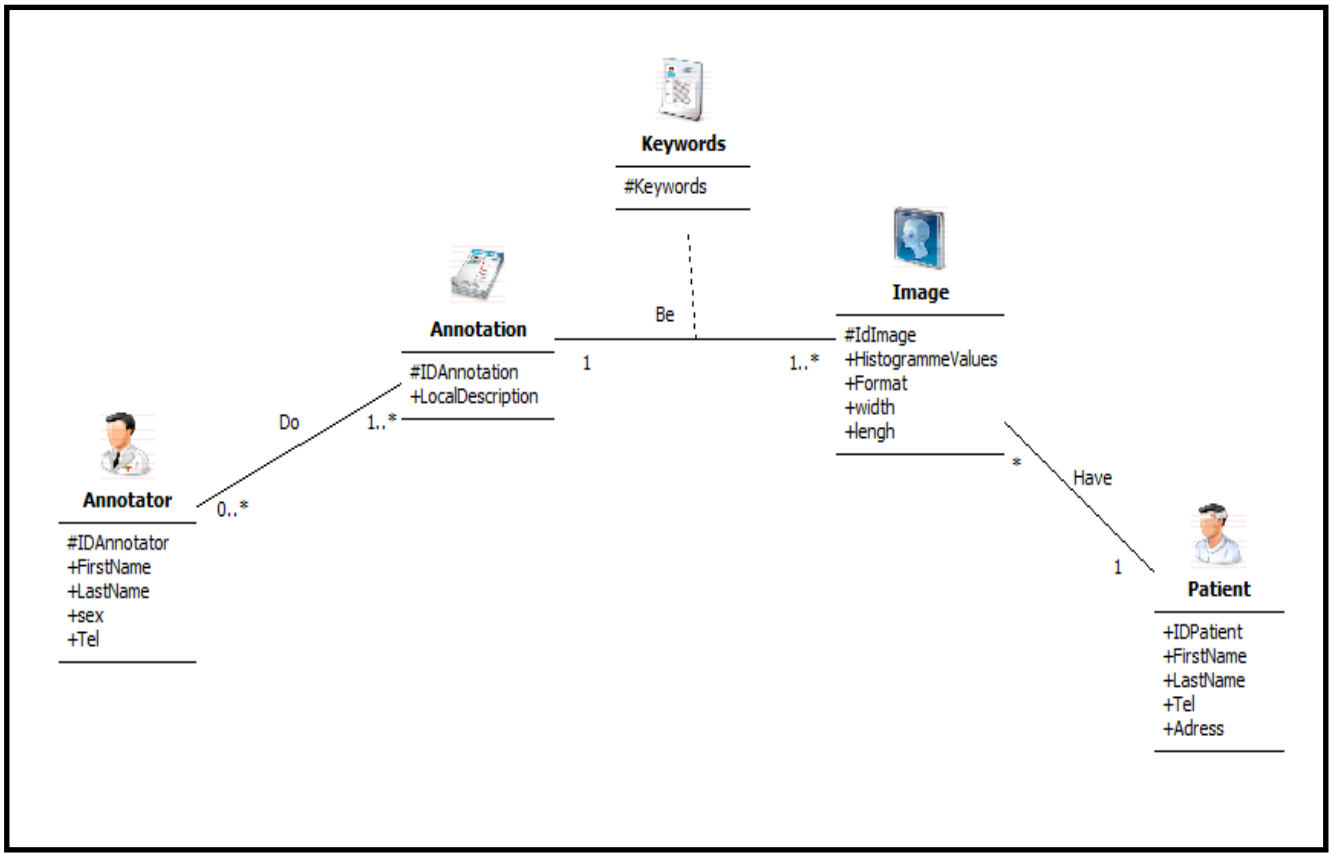

*Fig 5. The class diagram*

As defined in *Annotation Profile*, stereotypes can also be applied to relationships. Add a class to model and an element Association between them. To apply a stereotype to an element **Association** proceed the same way as described above for elements of type **Class**. Note that, now the only available stereotypes are those previously defined as being of type **UML Object** in the profile.

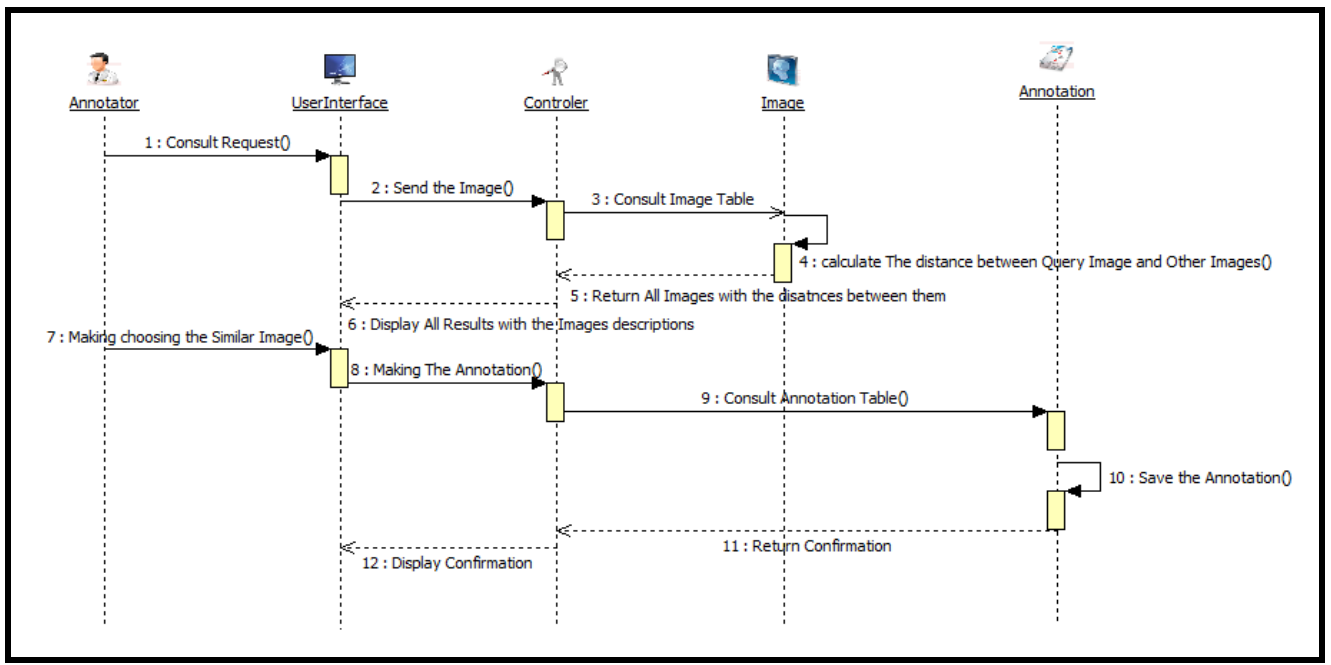

**Fig 5. The sequence diagram**

Remembering that *StarUML* only allows adding and viewing only one stereotype per element, and also that this tool does not support the definition of constraints on OCL language for profiles. Therefore, to verify the validity and consistency of the model is the responsibility of the designer.

## 6. CONCLUSION

Our In this paper, we described our profile named Medical Image Annotation. This profile contains the Class Diagram and sequence diagram which gives a conceptual representation of the annotation of medical image by specifying relationships between different entities of the diagram. We described the realization of the Annotation profile. To estimate our approach we ended this paper with an experimentation of the class diagram and sequence diagram. We propose as future work to represent a model of the component diagram that is based on an UML profile for the physical level.

## 7. REFERENCES

[1] L. K. Barnard, P. Duygulu, D. Forsyth, N. Freitas, D. Blei, and M. Jordan, "Matching words and pictures", JMLR, 2003.

[2] A. Farhadi, I. Endres, D. Hoiem, and D. Forsyth, "Describing objects by their attributes" CVPR, 2009.

[3] X. Wang, L. Zhang, M. Liu, Y. Li, and W. Ma," image search to annotation on billions of web photos" CVPR, 2010.

[4] J. Weston, S. Bengio, and N. Usunier, "Large scale image annotation: Learning to rank with joint wordimage embeddings", Machine Learning Journal, 2010.

[5] Igor Francisco Areias Amaral, "Content-Based Image Retrieval for Medical Applications ", October, 2010.

[6] B. Saida.**,** "Recherche d´images par contenu", 2007.

[7] T. Deselaers, D. Keysers, and H. Ney, "Flexible Image Retrieval Engine: Image CLEF 2004 Evaluation". In *Advances in Multilingual and Multimodal Information Retrieval, 5th Workshop of the Cross-Language Evaluation Forum, CLEF'04*, pages 688–698, 2004.

[8] H. Muller, W. Muller, S. Marchand-Maillet, S. March, T. Pun, and D. M. Squire. "Strategies for positive and negative relevance feedback in image retrieval", In *The 15th International Conference on Pattern Recognition, ICPR'00*, pages 1043–1046, 2000.

[9] W. D. Bidgood. "the SNOMED DICOM microglossary: controlled terminology resource for data interchange in biomedical imaging". DOLAP', 1998.

[10] S. J. Weston, S. Bengio, and N. Usunier, "Large scale image annotation: Learning to rank with joint wordimage embeddings", Machine Learning Journal, 2010.

[11] Stéphane Clinchant, Julien Ah-Pine, Gabriela Csurka," Semantic Combination of Textual and Visual Information in Multimedia Retrieval", 2011

[12] K Yiannis Gkfous, Anna Morou and Theodore Kalamboukis , "Combining Textual and Visual Information for Image Retrieval in the Medical Domain", 2011

[13] Brodeur, J., Badard, B.: Modeling with ISO 191xx standard. In: Shekhar, S.; Xiong, H. (Eds.). Encyclopedia of GIS. Springer-Verlag, pp. 691--700, 2008.

[14] Booch, G., Rumbaugh, J., Jacobson, I, "The Unified Modeling Language user guide. 2". ed. Addison-Wesley, Boston, 2005.

[15] Object Management Group, "Unified Modeling Language: Infrastructure. V. 2.1.2", 2007.

[16] Selic B. "The Pragmatics of Model-Driven Development." In: IEEE Software, vol. 20 nº 5, pp. 19–25, 2003.

[17] Pohjonen R., Kelly S." Domain-Specific Modeling." Dr. Dobb's Journal, 2002.

[18] Selic, B," A systematic approach to domain-specific language design using UML.", In: 10th IEEE Int. Symposium on Object and Component-Oriented Real-Time Distributed Computing (ISORC'07), pp. 2--9 (2007)

[19] Stempliuc, S. M., Lisboa F., J., Andrade, M. V. A., Borges, K. V. A," Extending the UML-GeoFrame data model for conceptual modeling of network applications", In: Int. Conf. on Enterprise Information Systems (ICEIS), Milão pp. 164--170 ,2009

[20] Bruck J., Hussey K.," Customizing UML: Which Technique is Right for You", IBM, 2007.

[21] Giachetti G., Marín B., Pastor O, "Using UML as a Domain-Specific Modeling Language: A Proposal for Automatic Generation of UML Profiles", 21st Conference on Advanced Information Systems Engineering (CAiSE'09). LNCS. Springer, pp. 110–124, 2009.

[22] Conallen, J,"Building Web Applications with UML", 2nd edn. AddisonWesley, 2002.

[23] Fons, J., Valderas, P., Ruiz, M., Rojas, G., Pastor, O, "OOWS: A Method to Develop Web Applications from Web Oriented Conceptual Models". Proceedings of the 7th World Multiconference on Systemics, Cybernetics and Informatics. Orlando, FL – USA , 2003

[24] Filipe Ribeiro Nalon and Karla A. de V. Borges, "A UML Profile for Conceptual Modeling in GIS Domain",2010

[25] M. Sharma, N. Rajpal and B.V.R.Reddy "Physical Data Warehouse Design using Neural Network". International Journal of Computer Applications 1(3):86–94, February 2010. Published By Foundation of Computer Science.

[26] S. Rizzi, Matteo Golfarelli,D.Maio "The Dimensional Fact Model: A Conceptual Model for Data Warehouses." International Journal of Cooperative Information Systems (IJC IS),7(2- 3):215-247,1998.

[27] E .Medina and S.L. Mora "A Web Oriented Approach to manage Multidimensional Models through XML Schemas and XSLT "EDBT 2002 Workshops, LNCS 2490, pp. 29–44, 2002. Springer-Verlag Berlin Heidelberg, 2002.

[28] S. Luján-Mora1, P. Vassiliadis and Juan Trujillo "Data Mapping Diagrams for Data Warehouse Design with UML"in Proceedings of 23 rd International Conference on Conceptual Modeling (ER 04),volume 3288 of LNCS,China,Springer, 2004.

[29] L. Mora and J. Trujilio "Physical Modeling of Data warehouses by using UML Component and Deployment Diagrams:Design and implementation issues. "Journal of Database Management 17(1), 2006.

[30] A. Gosain and S. Mann "Object Oriented Multidimensional Model for a Data Warehouse with Operators", International Journal of Database Theory and Application Vol. 3, No. 4, 2010.